\documentclass[10pt,twocolumn,letterpaper]{article}

\usepackage[pagenumbers]{cvpr} 

\usepackage{subcaption}
\usepackage{multirow}
\usepackage{enumitem, xspace}
\usepackage{listings}
\usepackage{xcolor}
\usepackage{amssymb}   
\usepackage{booktabs}
\usepackage{tabularx}
\usepackage{array}
\usepackage{microtype}

\definecolor{cfg-bg}{HTML}{FAFAFA}
\definecolor{cfg-nt}{HTML}{7C3AED}       
\definecolor{cfg-t}{HTML}{0E7490}        
\definecolor{cfg-op}{HTML}{B91C1C}       
\definecolor{cfg-comment}{HTML}{9CA3AF}  
\definecolor{cfg-str}{HTML}{047857}      
\definecolor{cfg-frame}{HTML}{E5E7EB}

\lstdefinestyle{cfgstyle}{
  backgroundcolor=\color{cfg-bg},
  basicstyle=\ttfamily\small\linespread{1.08}\selectfont,
  keywordstyle=\color{cfg-nt}\bfseries,
  keywordstyle=[2]\color{cfg-t}\bfseries,
  commentstyle=\color{cfg-comment}\itshape,
  stringstyle=\color{cfg-str},
  morekeywords={Plugin,MethodSpec,Pipeline,DataMgr,DataParser,
    Model,Optimizers,BaseOpt,CustomOpt,Viewer,Package,
    EntryPoints,Modification,ExtPoint},
  morekeywords=[2]{TrainerConfig,VanillaPipelineConfig,
    FullImageDMConfig,DepthDMConfig,FeatureDMConfig,CustomDMConfig,
    NerfstudioDPConfig,GSTemplateModelConfig,GSTemplateModel,
    SplatfactoModel,AdamOptimizerConfig,ViewerConfig,
    MethodSpecification,ExponentialDecayScheduler,
    populate__modules,split__gaussians,compute__custom__losses,
    get__outputs,get__gaussian__param__groups,get__param__groups,
    get__outputs__for__camera,load__state__dict,get__metrics__dict,
    DepthDataManager,means,features__dc,features__rest,
    opacities,scales,quats,camera__opt,
    pyproject.toml,configs.py,model.py,losses.py,rendering.py,
    datamanagers.py,methods,
    xys,radii,gauss__params,batch,info},
  morecomment=[l]{//},
  frame=single,
  rulecolor=\color{cfg-frame},
  framesep=5pt,
  xleftmargin=5pt,
  xrightmargin=3pt,
  aboveskip=8pt,
  belowskip=5pt,
  breaklines=true,
  columns=fullflexible,
  keepspaces=true,
  showstringspaces=false,
  tabsize=4,
  literate=
    {::=}{{{\color{cfg-op}\bfseries::=}}}{3}
    {|}{{{\color{cfg-op}\bfseries\textbar}}}{1}
    {*}{{{\color{cfg-op}\bfseries*}}}{1}
    {+}{{{\color{cfg-op}\bfseries+}}}{1}
    {<}{{{\color{cfg-nt}$\langle$}}}{1}
    {>}{{{\color{cfg-nt}$\rangle$}}}{1}
    ,
}

\lstdefinestyle{cfgstyle-fig}{
  style=cfgstyle,
  basicstyle=\ttfamily\fontsize{6.8pt}{8.6pt}\selectfont,
  framesep=3.5pt,
  xleftmargin=3pt,
  xrightmargin=2pt,
  aboveskip=4pt,
  belowskip=2pt,
}

\usepackage{tabularx}  
\usepackage{graphicx}
\usepackage{booktabs}

\usepackage[accsupp]{axessibility}  

\newcommand{\splatify}{\textsc{Splatify}}
\newcommand{\splatifybench}{\textsc{SplatifyBench}}

\definecolor{cvprblue}{rgb}{0.21,0.49,0.74}
\usepackage[pagebackref,breaklinks,colorlinks,allcolors=cvprblue]{hyperref}

\def\paperID{*****} 
\def\confName{3DV\xspace}
\def\confYear{2027\xspace}

\title{SPLATIFY: Reproduce, Discover, Innovate! \\ From Papers and Ideas to Trainable 3DGS Code}

\author{Seemandhar Jain \quad Keshav Gupta \quad Manmohan Chandraker\\
University of California, San Diego\\
{\tt\small \{sejain, keg019, mkchandraker\}@ucsd.edu}
}

\usepackage{cuted}
\usepackage{caption}
\begin{document}
\maketitle

\begin{strip}
  \centering
  \includegraphics[width=0.70\textwidth]{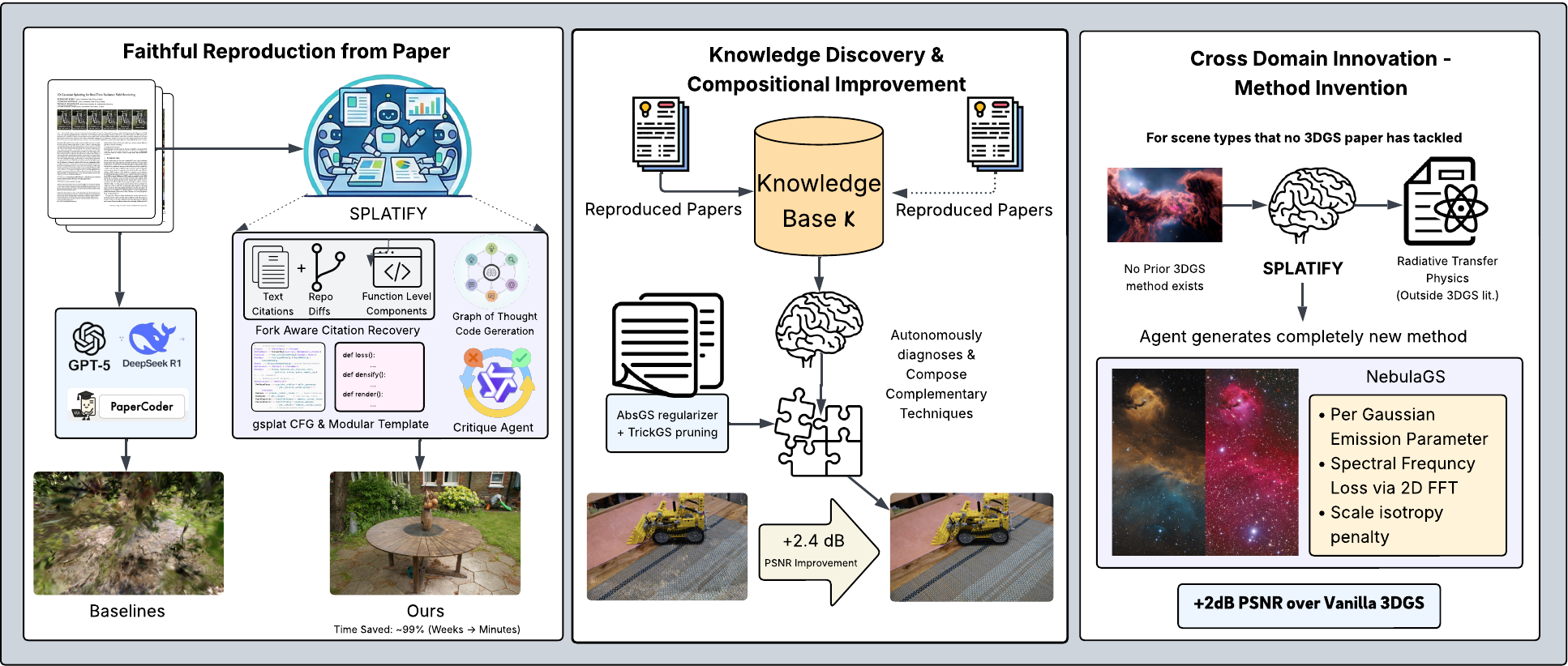}
  \captionof{figure}{
    \textbf{\splatify: Reproduce, Discover, Innovate.}
    Given a 3DGS paper, \splatify\ produces trainable gsplat code in minutes
    existing approaches cannot (Level 1), autonomously composes complementary
    techniques for up to $+2.4\,\mathrm{dB}$ PSNR improvement for existing
    literature (Level 2), and invents entirely new methods by retrieving
    physical priors from outside the 3DGS literature (Level 3).
  }
  \label{teaser}
\end{strip}

\begin{abstract}
The rapid growth of 3D Gaussian Splatting (3DGS) research demands significant effort to reimplement papers before building on them. We introduce \textsc{Splatify}, a multi-agent framework that converts 3DGS papers into trainable gsplat-based implementations, where generic paper-to-code methods and frontier models fail. \textsc{Splatify} achieves this through five innovations: (1) A \textbf{context-free grammar for gsplat} over a modular method template with extension points for losses, densification, rendering, and optimization, constraining synthesis so generated code satisfies gsplat's architectural invariants by construction. (2) \textbf{Architectural elements for faithful reproduction}: fork-aware citation recovery retrieving component-level code at function-level granularity, Graph-of-Thought synthesis in topological dependency order, RAG-guided in-context example selection from over 20 verified implementations, and visual feedback combining PSNR-guided regeneration, Gaussian-level structural checks, and VLM-driven patching. (3) \textbf{Knowledge-driven compositional improvement} that autonomously finds weaknesses and composes complementary regularizers, losses, and densification strategies to improve upon original results. (4) \textbf{Interdisciplinary method discovery} where agents retrieve physical priors from outside the 3DGS literature and compose them with rendering knowledge to produce methods for previously unaddressed scene types. (5) \textbf{\textsc{Splatify-Bench}}, an evaluation framework across 30 diverse 3DGS papers. On papers without public code, \textsc{Splatify} matches expert implementations while reducing development time from weeks to minutes, and through compositional discovery further improves PSNR by up to 2.4\,dB. We additionally demonstrate novel methods for volumetric nebula rendering and other scientific domains, synthesized entirely by \textsc{Splatify}. Code, data, and all implementations will be publicly released.
\end{abstract}

\vspace{-0.3cm}
\section{Introduction}
\label{sec:intro}

\textbf{3D Gaussian Splatting has triggered a fast-growing sub-field} of neural rendering, yet the community publishes knowledge far faster than it can absorb. Since Kerbl et al.~\cite{kerbl20233dgs} introduced 3DGS in 2023, follow-up papers have grown to over 1,500 in under three years~\cite{chen2024survey3dgs, fei20243dgsnewera}, with tracking repositories recording roughly three new arXiv submissions per day~\cite{kulhanek2024nerfbaselines}. For a researcher who wants to \emph{use} a recently published method, however, the path from paper to working code remains prohibitively slow.

\textbf{Existing infrastructure cannot keep pace with this growth.}  While over 40\% of tracked 3DGS papers release no code~\cite{awesome3dgs}, others publish isolated forks with incompatible assumptions about attribute layouts, density control, rendering conventions~\cite{kulhanek2024nerfbaselines}, and CUDA kernels. Building on a recent method typically demands weeks of manual reimplementation and debugging. Generic pipelines like Paper2Code~\cite{seo2025paper2code} and AutoP2C~\cite{lin2025autop2c} produce code that do not compile against any rasterization backend, and frontier models generate Python that collapses at training time for lack of domain knowledge about Gaussian attribute coupling or adaptive density control. No existing paper-to-code system, in any domain, handles 3DGS's fork-dominated landscape, provides Gaussian-level structural verification, or moves beyond reproduction to compositional improvement.

\textbf{Even if reproduction were perfectly solved, the deeper bottleneck would persist: the knowledge across 2k+ papers is neither discoverable nor composable.} There is no efficient way to determine whether combining a depth distortion loss from 2DGS~\cite{huang20242dgs} with stochastic splitting from 3DGS-MCMC~\cite{kheradmand20243dgsmcmc} and an opacity regularizer from AbsGS~\cite{ye2024absgs} would outperform any of them individually. With hundreds of published techniques, the number of valid compositions is enormous, and manual exploration covers a negligible fraction. Entire application domains where Gaussians are a natural geometric fit, from optically thin nebulae to fluorescence microscopy, remain untouched because the engineering barrier is too high. The field needs infrastructure for \emph{discovering}, \emph{composing}, and \emph{extending} the knowledge its papers contain.

\textbf{We introduce \textsc{Splatify} (Fig.~\ref{teaser})}, a multi-agent framework operating at three levels: faithful reproduction of 3DGS papers as trainable gsplat~\cite{ye2025gsplat} plugins, compositional improvement that autonomously combines complementary techniques, and cross-domain method invention for unaddressed scene types. Its design rests on five contributions:
\begin{enumerate}[leftmargin=*,itemsep=1pt,parsep=0pt,topsep=3pt]
\item \textbf{A context-free grammar for gsplat plugins} over a modular template with nine extension points covering the full 3DGS modification space, ensuring generated code satisfies architectural invariants by construction (Sec.~\ref{sec:method_template}).
\item \textbf{Faithful reproduction} extending the multi-stage pipeline of~\cite{jain2026nerfifymultiagentframeworkturning} with fork-aware citation recovery at function-level granularity, Gaussian-specific smoke tests, and PSNR training-curve monitoring (Sec.~\ref{methodproposed}).
\item \textbf{Compositional improvement} through autonomous diagnosis of weaknesses in reproduced methods and composition of complementary regularizers, losses, and densification strategies from the knowledge base, yielding up to $+$2.4\,dB PSNR gain with no human guidance (Sec.~\ref{sec:knowledge_discovery}).
\item \textbf{Cross-domain method invention}, where agents retrieve physical priors from outside the 3DGS literature and synthesize physics-informed modifications for unaddressed domains such as volumetric nebula rendering and thin-film bubble iridescence (Sec.~\ref{sec:knowledge_discovery}).
\item \textbf{\textsc{Splatify}-Bench}, a 30-paper benchmark stratified by code availability and technical complexity (Sec.~\ref{sec:splatify-bench}).
\end{enumerate}
On papers without public code, \textsc{Splatify} matches expert implementations within $\pm$0.5\,dB PSNR and $\pm$0.02 SSIM while existing baselines fail to produce trainable code. Cross-domain invention produces working methods for scene types no prior paper has addressed, entirely without human guidance. \textsc{Splatify}, \textsc{Splatify-Bench}, all implementations, and the code will be publicly released.
\vspace{-0.3cm}
\section{Related Work}
\label{sec:related_work}
\noindent\textbf{The 3DGS Reproducibility Problem.}
3D Gaussian Splatting~\cite{kerbl20233dgs} represents scenes as anisotropic Gaussians rasterized through a differentiable tile-based renderer. The field has grown to over a thousand papers~\cite{chen2024survey3dgs,fei20243dgsnewera} spanning anti-aliasing~\cite{yu2024mipsplatting}, geometry~\cite{huang20242dgs}, densification~\cite{kheradmand20243dgsmcmc}, and compression~\cite{lee2024compact3dgs,fan2023lightgaussian}. We target gsplat~\cite{ye2025gsplat} integrated with Nerfstudio~\cite{tancik2023nerfstudio} as the compilation target. A fundamental barrier is the fragmented implementation landscape: most papers fork the original INRIA codebase into one-off repositories that resist comparison or reuse, demanding deeper domain awareness than any existing system provides.

\noindent\textbf{Automated Research Reproduction.}
Paper2Code~\cite{seo2025paper2code} and AutoP2C~\cite{lin2025autop2c} use staged parsing but produce non-trainable outputs for vision methods; AutoReproduce~\cite{zhao2025autoreproduce} and RePro~\cite{zhou2025repro} improve fidelity through citation lineage extraction yet remain domain-agnostic. PaperBench~\cite{starace2025paperbench} quantifies the gap: the strongest agent scores 21\% against 41.4\% for human researchers on ICML papers. Multi-agent frameworks including MetaGPT~\cite{hong2024metagpt}, ChatDev~\cite{qian2024chatdev}, AutoGen~\cite{wu2024autogen}, and repository-level agents~\cite{wang2024codeact,yang2024sweagent} advance role-specialized collaboration but lack scientific rendering-specific reasoning. 
The closest prior work is \textsc{Nerfify}~\cite{jain2026nerfifymultiagentframeworkturning}, which introduces grammar-constrained generation, compositional citation recovery, and visual feedback for NeRF-to-Nerfstudio synthesis. However, \textsc{Nerfify} is limited to NeRFs, and 3DGS methods alter tightly coupled components across fragmented repository forks, unlike NeRF systems that modify relatively isolated neural modules. \textsc{Nerfify} cannot handle this fork-dominated ecosystem and lacks Gaussian-level verification. SPLATIFY adds fork-aware recovery via parent-repository diffs and Gaussian structural checks with PSNR curve monitoring, and goes beyond reproduction: it composes techniques to surpass the reproduced result and retrieves physical priors from outside the field for unaddressed scene types.

\noindent\textbf{Structured Reasoning and Iterative Code Refinement.}
Constraining LLM outputs with formal structure improves code correctness, as shown by pushdown-automaton decoding~\cite{dong2023codep}, grammar prompting~\cite{wang2023grammarprompting}, and CFG-based visual program synthesis~\cite{zhang2024scenelanguage}. Iterative refinement through execution feedback is equally standard: Graph of Thoughts~\cite{besta2024got} and Tree of Thoughts~\cite{yao2023tot} organize reasoning as searchable structures, while Self-Debugging~\cite{chen2024selfdebugging}, Reflexion~\cite{shinn2023reflexion}, and Self-Refine~\cite{madaan2023selfrefine} enable models to critique and repair their own outputs. \textsc{Splatify} combines both: a gsplat CFG guarantees architectural validity, while PSNR training-curve analysis and VLM-guided patching provide closed-loop feedback to converge on expert-quality implementations.

\section{Proposed Approach}

\begin{figure}[t]
    \centering
    \includegraphics[width=0.9\linewidth]{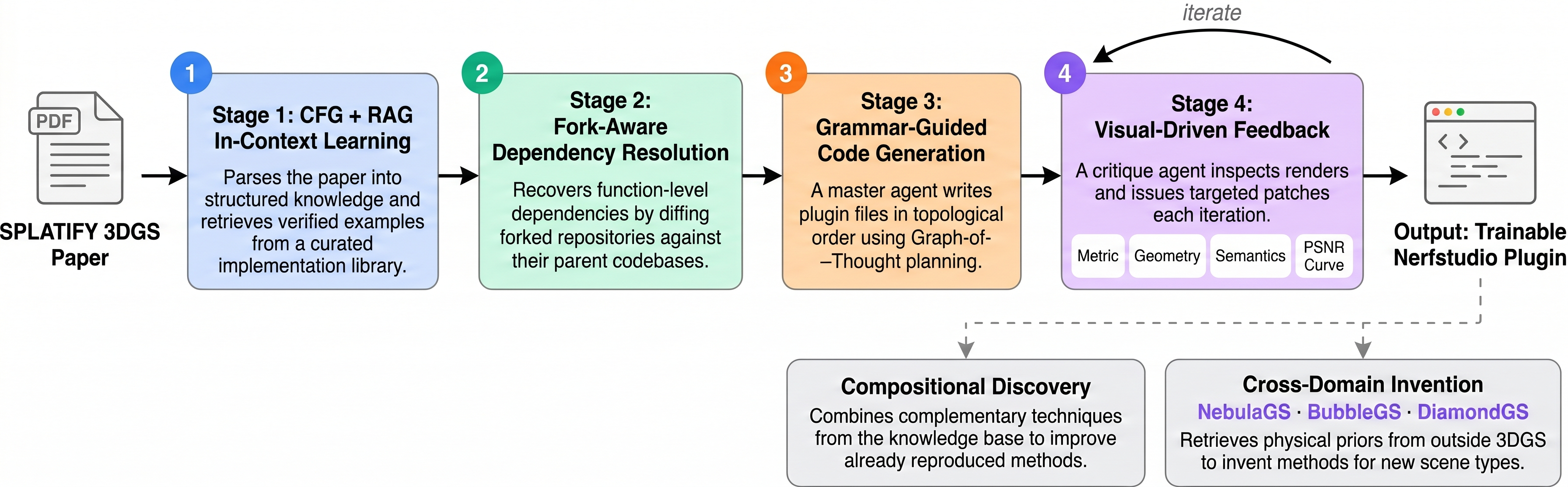}
\caption{\textbf{Simplified SPLATIFY pipeline.}
A high-level view of reproduction, refinement, and interdisciplinary method discovery.
Fig.~\ref{fig:overview} shows the detailed architecture.}
    \label{fig:simplified-overview}
\end{figure}

\begin{figure*}[t]
  \centering
  \includegraphics[width=\textwidth]{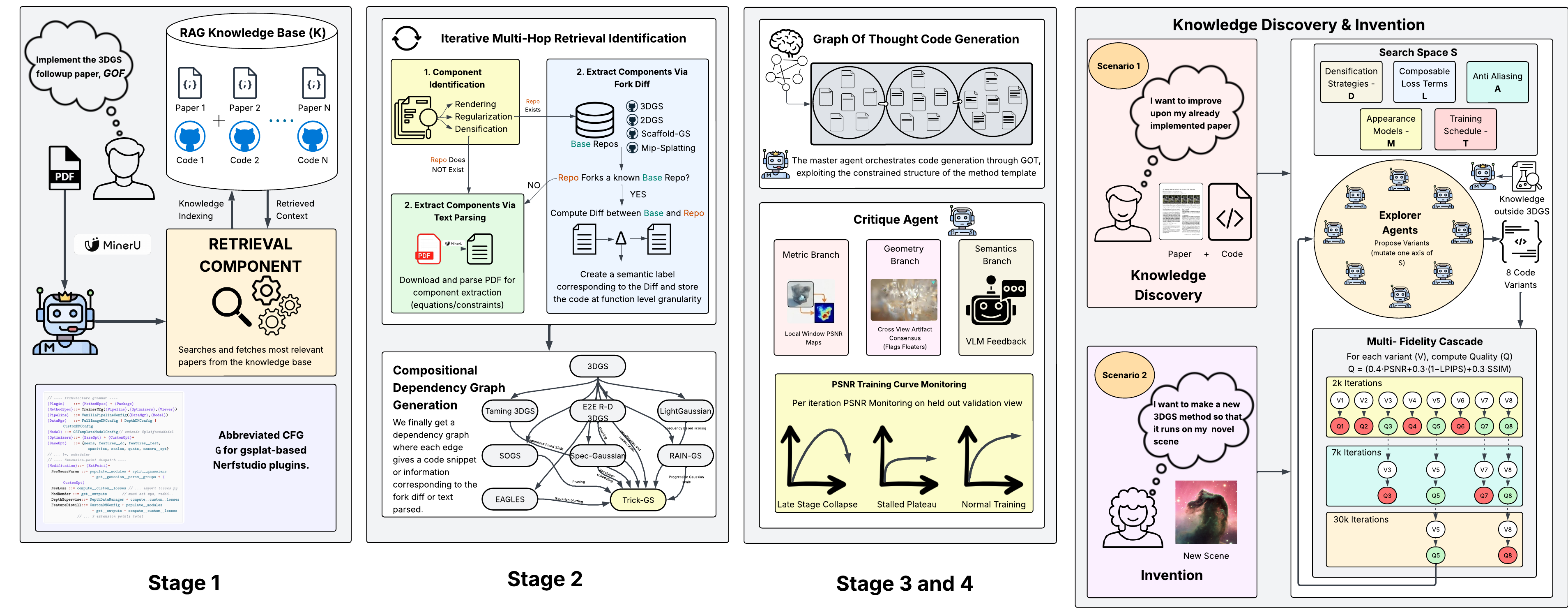}
  \caption{\textbf{\splatify\ overview.} \emph{Reproduction} (left): a parsing 
  agent structures papers into a CFG-grounded knowledge base; fork-aware dependency 
  resolution recovers components via citation graphs and repository diffs; GoT 
  synthesis generates code in topological order with PSNR-guided and VLM-driven 
  refinement. \emph{Knowledge discovery and invention} (right): the populated 
  knowledge base enables compositional improvement, where agents compose techniques 
  to surpass original results, and interdisciplinary method creation, where physical priors from outside 3DGS literature are retrieved to address novel scene types.}
  \label{fig:overview}
\end{figure*}

\subsection{Problem Statement}
We consider three objectives of increasing scope: \emph{faithful reproduction},
\emph{compositional improvement}, and \emph{interdisciplinary method discovery}.

\noindent\textbf{Faithful Reproduction.}
Given a 3DGS paper $\mathcal{P}$, we synthesize an executable repository
$\mathcal{C} = (F, G)$, where $F = \{f_1, \ldots, f_n\}$ is a set of source files and
$G = \textsc{BuildRepoDAG}(F)$ encodes their import and dataflow dependencies.
A parsing pipeline extracts structured information
$\mathcal{E}(\mathcal{P}) = \langle T, I, Q, B \rangle$ (text, figures, equations, bibliography) and passes it to our multi-agent system
$\mathcal{A}: \bigl(\mathcal{E}(\mathcal{P});\,\mathcal{R}\bigr) \mapsto \mathcal{C}$.
The resource tuple $\mathcal{R} = (\mathcal{K}, \mathcal{W}, \mathcal{X})$ comprises a CFG-grounded gsplat knowledge base $\mathcal{K}$ (Sec.~\ref{sec:method_template}),
web resources $\mathcal{W}$ for cited papers and fork-level diffs, and a library of verified implementations $\mathcal{X}$ serving as in-context examples. Reproduction succeeds when $\mathcal{C}$ compiles against gsplat, trains to convergence, and achieves rendering quality within $\epsilon = 0.5$\,dB PSNR of reported metrics while correctly implementing every novel component in $\mathcal{P}$.

\noindent\textbf{Compositional Improvement.}
Many 3DGS methods address a single weakness while remaining vulnerable to others.
Given a faithful reproduction $\mathcal{C}_0$ and a library of verified techniques
$\mathcal{K}_{\mathrm{tech}} = \{t_1,.., t_k\} \subset \mathcal{K}$, the agent uses
visual feedback to find weaknesses, queries $\mathcal{K}$ for complementary remedies,
and validates that
$\mathcal{C}^{+}=\textsc{Compose}(\mathcal{C}_0,\, S \subseteq \mathcal{K}_{\mathrm{tech}})$
remains a valid grammar derivation and improves quality on held-out views.
The agent receives no human guidance; it must independently diagnose, retrieve, and verify.

\noindent\textbf{Interdisciplinary Method Discovery.}
The furthest-reaching objective targets scene types no published 3DGS paper has addressed,
such as volumetric astronomical phenomena or fluorescence microscopy. Given a target domain
$\mathcal{D}$ and a multi-view training set, the agent retrieves physical priors
$\mathcal{W}_{\mathrm{phys}} \subset \mathcal{W}$ from literature outside 3DGS, identifies
which splatting assumptions are violated, and synthesizes physics-informed modifications
within gsplat:
$\mathcal{C}_{\mathcal{D}} = \mathcal{A}\bigl(\mathcal{D},\,\mathcal{W}_{\mathrm{phys}};\,
\mathcal{K},\,\mathcal{G}\bigr)$.
Success is measured by improvement over vanilla 3DGS, cross-view consistency, and physical
plausibility. This moves beyond translating existing knowledge toward generating new knowledge
by bridging disciplinary boundaries.

\subsection{CFG-Constrained gsplat Method Template}
\label{sec:method_template}
Unconstrained LLM generation can produce code that looks plausible but violates framework invariants in ways that surface only at training time: dead optimizer entries, missing density-control bookkeeping, losses wired into the wrong module. Our solution is to formalize gsplat-based Nerfstudio plugins as a CFG $\mathcal{G} = (\Sigma, V, S, P)$ and provide agents with a concrete method template instantiating this grammar as runnable code. This makes synthesis a constrained task where agents decide \emph{which} extension points a paper's contributions map to and \emph{what} code to place there, while the structural skeleton is fixed by construction.

\noindent\textbf{Template design.}
By analysis of over 40 3DGS papers spanning anti-aliasing~\cite{yu2024mipsplatting}, geometry~\cite{huang20242dgs}, densification~\cite{kheradmand20243dgsmcmc}, depth supervision, and feature distillation, we identify \textbf{nine extension points} covering the full modification space (listed in the supplementary). The template's core class \texttt{GSTemplateModel} inherits tile-based gsplat rasterization, adaptive density control, SH schedules, camera optimization, and the standard optimizer set. Agents subclass it and override only relevant hooks: \texttt{compute\_custom\_losses()} for a new regularizer, or \texttt{populate\_modules()} plus \texttt{split\_gaussians()} for a new per-Gaussian attribute.

\noindent\textbf{The grammar.}
Terminals $\Sigma$ are framework classes, non-terminals $V$ are abstract components, $S{=}\langle\textsc{Plugin}\rangle$ is the start symbol, and productions $P$ encode valid compositions. The grammar serves two roles: it is included in every agent's system prompt as a structural specification, and it defines the verification contracts used during GoT synthesis (Sec.~\ref{methodproposed}), where the master agent checks after each generated file that the repository remains a valid derivation of $\mathcal{G}$. An abbreviated version is shown in Fig.~\ref{fig:overview} Stage 1; the full grammar is in the supplementary.

\subsection{Proposed Multi-Stage Framework}\label{methodproposed}
Fig.~\ref{fig:simplified-overview}
provides a simplified view of the overall SPLATIFY pipeline, while
Fig.~\ref{fig:overview} shows the complete system architecture. We design a multi-stage framework similar to \cite{jain2026nerfifymultiagentframeworkturning}, but with several extensions and 3DGS-specific components, which are described below. 

\subsubsection{Stage 1: RAG-Guided In-Context Learning and CFG}
\splatify\ centers on the knowledge base $\mathcal{K}$ and a library of verified implementations $\mathcal{X} = \{(\mathcal{P}_i, \mathcal{C}_i)\}_{i=1}^{|\mathcal{X}|}$, seeded with 20 manually validated papers. The agent first converts $\mathcal{P}$ to structured markdown via MinerU~\cite{niu2025mineru25decoupledvisionlanguagemodel}, stripping non-technical sections while preserving equations, pseudocode, diagrams, and citation relationships. For each new target, a retrieval agent identifies the relevant extension points and selects the most similar subset of $\mathcal{X}$ by extension-point overlap and technique type, so a paper proposing a regularization loss retrieves other regularization implementations. Any paper \splatify\ successfully implements is automatically added back to $\mathcal{X}$, progressively broadening coverage.

\subsubsection{Stage 2: Fork-Aware Compositional Dependency Resolution}
3DGS papers are deeply compositional: a single method routinely assembles rendering, regularization, densification, and post-processing components from many prior works. Unlike NeRF methods, which ship as standalone packages, roughly 70\% of 3DGS papers release code as forks of the INRIA repository~\cite{kerbl20233dgs}. A system that retrieves only text-level references misses these code-level dependencies entirely.

\noindent {\em Fork-aware recovery.}
The agent builds a dependency graph via iterative multi-hop retrieval, identifying borrowed components, recursively resolving transitive dependencies, and terminating when all interface contracts are satisfied. The key departure from prior work is at the extraction step: for each cited paper, the agent checks whether its repository forks a known base (3DGS, Mip-Splatting, 2DGS, Scaffold-GS), computes a structured diff against the parent, and maps each changed function to a semantic label (e.g., \texttt{loss/depth\_distortion}), yielding exact function-level contributions rather than noisy text-level citations. When no repository exists, the agent falls back to PDF-level extraction. Transitive fork chains are resolved by chaining diffs to recover the full modification stack.

\noindent {\em Running example.}
Figure~\ref{fig:overview} Stage 1 shows Trick-GS~\cite{armagan2025trickgs}, which assembles 8 techniques from 7 papers: volume masking from Self-Organizing Gaussians~\cite{morgenstern2024compact}, significance pruning from LightGaussian~\cite{fan2023lightgaussian}, SH band masking from E2E-RD~\cite{wang2024end}, progressive training from EAGLES~\cite{girish2024eagles} and Spec-Gaussian~\cite{yang2024specgaussian}, progressive scales from RAIN-GS~\cite{jung2024raings}, and accelerated rasterization from Taming 3DGS~\cite{mallick2024taming}. Several of these fork INRIA 3DGS. Our agent diffs each fork against its parent (Fig.~\ref{fig:overview}), tags changed functions with semantic labels, and chains results across transitive edges, expanding the graph to 12 papers. Such depth is common: GaussianPro~\cite{cheng2024gaussianpro} spans 8 dependencies, while Scaffold-GS~\cite{lu2024scaffoldgs} and PGSR~\cite{chen2024pgsr} each span 7.

\subsubsection{Stage 3: Grammar-Guided Repository Generation}
The master agent orchestrates a Graph-of-Thought (GoT) code generation approach exploiting the constrained structure of our method template. Since every plugin subclasses \texttt{GSTemplateModel} and inherits the full splatfacto training loop, rasterization, density control, and optimizer wiring, synthesis reduces to identifying extension points to which a paper's contributions map and filling in only the relevant hooks. The agent classifies contributions into one or more of the nine extension points defined by $\mathcal{G}$, verifies the selected combination before writing code, then generates files in topological order with DAG construction, interface freezing, implementation, and smoke testing. Most repositories pass within two iterations.

\subsubsection{Stage 4: Visual-Driven Feedback}
After the system produces trainable code, we perform smoke training for 3k iterations and dispatch rendered images to a critique agent operating on four branches (Fig.~\ref{fig:overview}): a \emph{metric branch} computing local-window PSNR and SSIM error maps, a \emph{geometry branch} flagging floaters and ghosting via cross-view artifact consensus, a \emph{semantics branch} using a VLM to produce structured diagnoses with candidate patches, and a \emph{training-curve branch} unique to \splatify. The last branch exploits a key difference from NeRF: because 3DGS trains in minutes, the agent can monitor the full PSNR trajectory. 3DGS has a structurally distinct densification phase terminating around 15k iterations, after which errors in pruning or loss schedules manifest as failures invisible during short smoke runs. The agent flags two modes: a \emph{late-stage collapse} (PSNR drops sharply post-15k) and a \emph{stalled plateau} (PSNR improves by less than $0.1$\,dB over any $5$k-iteration window after warm-up). In both cases the agent localizes the fault to a specific extension point and issues a targeted patch. The loop continues until no further feedback is produced, the iteration budget is exhausted, or the paper's reported PSNR is matched.

\subsection{Knowledge Discovery via Evolutionary Technique Composition}
\label{sec:knowledge_discovery}
Beyond faithful reproduction, \splatify\ operates in two modes (Fig.~\ref{fig:overview}, right): \emph{compositional discovery}, which improves a synthesized implementation by composing complementary techniques from the 3DGS literature, and \emph{cross-domain invention}, which retrieves physical priors from outside the field to address scene types no prior paper has tackled.

\noindent\textbf{Technique search space.} The gsplat extension points define a structured configuration $\mathcal{S} = (\mathcal{D}, \mathcal{L}, \mathcal{A}, \mathcal{M}, \mathcal{T})$: mutually exclusive densification strategies $\mathcal{D}$ (ADC, AbsGrad, MCMC), composable loss terms $\mathcal{L} = 2^{\{\ell_1,\dots,\ell_n\}}$ (depth, normal, flatness, distortion), anti-aliasing options $\mathcal{A}$, appearance models $\mathcal{M}$, and continuous training-schedule parameters $\mathcal{T}$. Losses compose additively; densification strategies are mutually exclusive as they govern the same split-and-prune logic. For cross-domain invention, $\mathcal{L}$ is augmented with physics-informed losses retrieved from outside the 3DGS literature.

\noindent\textbf{Evolutionary search.} Inspired by FunSearch~\cite{FunSearch2023}, we maintain $K{=}8$ islands each managed by an explorer agent. Each round proceeds in three steps (Fig.~\ref{fig:overview}): (1)~each explorer proposes a variant by mutating one axis of $\mathcal{S}$ or composing an additional loss or strategy from $\mathcal{K}$ when a diagnosed weakness suggests it; (2)~the Stage~3 pipeline converts each proposal into runnable gsplat code, evaluated via a multi-fidelity cascade ($2\text{k}{\to}7\text{k}{\to}30\text{k}$ iterations, eliminating the bottom half at each stage) to reduce per-round compute; (3) agents are ranked by Pareto sorting over normalized
$0.4{\cdot}\text{PSNR}{+}0.3{\cdot}(1{-}\text{LPIPS}){+}0.3{\cdot}\text{SSIM}$
on subset of training views. The loop terminates when no agent improves for three consecutive rounds or the round budget is exhausted. For discovery, the search space is restricted to techniques already in $\mathcal{K}$; for invention, the agents additionally query $\mathcal{W}_{\mathrm{phys}}$ to retrieve domain-specific priors and propose physics-informed losses not present in any existing 3DGS paper. We demonstrate both modes in Sec.~\ref{sec:discovered-improvements}: compositional discovery yields up to $+2.4$\,dB PSNR over the reproduction-only baseline, and cross-domain invention produces new methods for volumetric nebula rendering and thin-film bubble iridescence.
\section{Experiments}
\label{sec:experiments}

We evaluate \textsc{Splatify} on \textsc{Splatify-Bench}, a new benchmark for 3DGS paper-to-code synthesis, and present results on reproduction quality (Sec.~\ref{sec:comparison-expert}), novelty preservation (Sec.~\ref{sec:novelty-coverage}), knowledge discovery (Sec.~\ref{sec:knowledge-discovery}), and ablations.

\subsection{\textsc{Splatify-Bench}}
\label{sec:splatify-bench}

Existing paper-to-code benchmarks~\cite{starace2025paperbench,seo2025paper2code} evaluate general ML code generation and lack criteria specific to 3DGS, where implementations must coordinate Python-level training logic with a CUDA rasterization backend, manage heuristic adaptive density control, and navigate a fragmented ecosystem of forked repositories. \textsc{Splatify-Bench} is the first benchmark targeting this domain, comprising \textbf{30 papers} across three sets.

\noindent\textbf{Set~1: Never-Implemented Papers (10).} Papers with no publicly available code, paired with expert reference implementations we commissioned. Since no implementation exists online, LLMs cannot have encountered corresponding code during pretraining, ensuring uncontaminated evaluation.~\cite{nguyen2025dwtgs,huang2025decomposingdensification,pateux2025bogauss,farooq2025opti3dgs,yan2024msaa3dgs,cheng2025clodgs,nowak2025vod3dgs,yang2026tidigs,elrawy2025opacitygradientdrivendensitycontrol,taghipour2025svrgsspatiallyvariantregularization}

\noindent\textbf{Set~2: Papers with Public Code (10).} Papers with released repositories spanning densification strategies, regularization losses, and appearance modeling, all portable to gsplat, allowing direct comparison against original author code.\cite{kerbl20233dgs,zhang2024lp,deng2025naturalselection,lee2025microsplatting,navaneet2024compgs,grubert2025improving,chen2025dashgaussian,chung2024depth,han2024binocular,park2025dropgaussian}

\noindent\textbf{Set~3: Novelty Coverage Papers (10).} Papers chosen for the number and clarity of their distinct technical contributions, each with multiple verifiable novelty items (loss formulations, architectural modules, training schedules, Gaussian attribute modifications), used exclusively for the novelty analysis in Sec.~\ref{sec:novelty-coverage}.~\cite{armagan2025trickgs,huang20242dgs,jiang2024gaussianshader,kheradmand20243dgsmcmc,lee2024compact3dgs,lu2024scaffoldgs,yang2024specgaussian,yu2024gof,zhang2024fregs,zhang2024pixelgs}

\subsection{Baselines, Metrics, and Experimental Setup}
\label{sec:baselines}

\noindent\textbf{Baselines.} We compare against five systems: generic paper-to-code pipelines Paper2Code~\cite{seo2025paper2code} and AutoP2C~\cite{lin2025autop2c}; frontier LLMs such as GPT-5.2 ET, Gemini~3.1~Pro, and DeepSeek~3.2; and expert implementations by researchers with 3DGS experience (typically 1--2 weeks per paper). \textsc{Splatify} uses the full pipeline with GPT-5.2 as the base LLM for agents. All baselines were evaluated under the same setup: identical paper PDF and supplementary, prompts, and compute budget.

\noindent\textbf{Metrics.} We report PSNR (dB), SSIM, LPIPS (lower is better), and Gaussian count (\#GS, millions) as a memory proxy. \textsc{Splatify} completes in 10--30 minutes per paper. API cost analysis is in the supplementary.

\noindent\textbf{Datasets and protocol.} Our primary benchmark is Mip-NeRF~360~\cite{barron2022mipnerf360} (five outdoor, four indoor unbounded scenes). We follow the evaluation conventions of \cite{bagdasarian20253dgszip}. Every 8th image is held out for testing. We report on all nine scenes including the often-omitted \textit{flowers} and \textit{treehill}, and restrict Tanks and Temples to \textit{train} and \textit{truck}. All methods train for 30k iterations on identical data and test viewpoints on NVIDIA A6000 GPUs (48\,GB), averaged on multiple seeds. Per-scene breakdowns are in the supplementary.

\subsection{Comparison to Expert Implementations}
\label{sec:comparison-expert}

\subsubsection{Set 1: Never-Implemented Papers}
\label{sec:set1}
We first evaluate \splatify\ on papers for which no public implementation exists. This setting provides the strongest test of paper-driven synthesis capability: since no code for these methods has been released, the LLMs powering both our agents and the baselines cannot have encountered corresponding implementations during pretraining, ruling out data contamination. Expert developers required one to two weeks per paper to produce reference implementations, while \splatify\ generates comparable code in minutes.

\begin{table}[t]
  \caption{\textbf{\splatify\ vs.\ reported results and human implementations on Set~1.} All other baselines failed to produce trainable code for any paper. Full results in the supplementary.}
  \label{tab:comparison}
  \centering
  \resizebox{\linewidth}{!}{
  \begin{tabular}{@{}llcccccccccccc@{}}
    \toprule
    \textbf{Paper} & \textbf{Scene} & \multicolumn{4}{c}{\textbf{Reported}} & \multicolumn{4}{c}{\textbf{Expert Impl.}} & \multicolumn{4}{c}{\textbf{\splatify\ (Ours)}}  \\ 
    \cmidrule(lr){3-6} \cmidrule(lr){7-10} \cmidrule(lr){11-14}
     & & PSNR$\uparrow$ & SSIM$\uparrow$ & LPIPS$\downarrow$ & \#GS ($10^6$)  & PSNR$\uparrow$ & SSIM$\uparrow$ & LPIPS$\downarrow$ & \#GS ($10^6$)  & PSNR$\uparrow$ & SSIM$\uparrow$ & LPIPS$\downarrow$ & \#GS ($10^6$)  \\ 
    \midrule
    DWTGS~\cite{nguyen2025dwtgs}                                       & mipnerf360 & 19.75 & 0.590 & 0.343 & --    & 17.92 & 0.508 & 0.471 & 1.21 & 18.68 & 0.510 & 0.374 & 1.641 \\
    Dec. Densification~\cite{huang2025decomposingdensification} & mipnerf360 & 27.79 & 0.826 & 0.214 & 1.469 & 26.85 & 0.792 & 0.286 & 1.44 & 27.17 & 0.820 & 0.261 & 0.650 \\
    BOGausS~\cite{pateux2025bogauss}                                   & mipnerf360 & 27.54 & 0.804 & 0.251 & 0.32  & 27.25 & 0.801 & 0.276 & 0.81 & 27.34 & 0.809 & 0.263 & 0.440 \\
    Opti3DGS~\cite{farooq2025opti3dgs}                                 & mipnerf360 & 27.09 & 0.809 & 0.237 & 1.327 & 27.51 & 0.816 & 0.221 & 6.18 & 27.71 & 0.841 & 0.209 & 1.406 \\
    CLoD-GS~\cite{cheng2025clodgs}                                     & Truck      & 25.29 & 0.880 & 0.148 & 1.46  & 24.83 & 0.854 & 0.227 & 0.40 & 25.48 & 0.873 & 0.128 & 0.384 \\
    \bottomrule
  \end{tabular}
  }
\end{table}
\begin{figure}[t]
  \centering
  \includegraphics[width=\columnwidth]{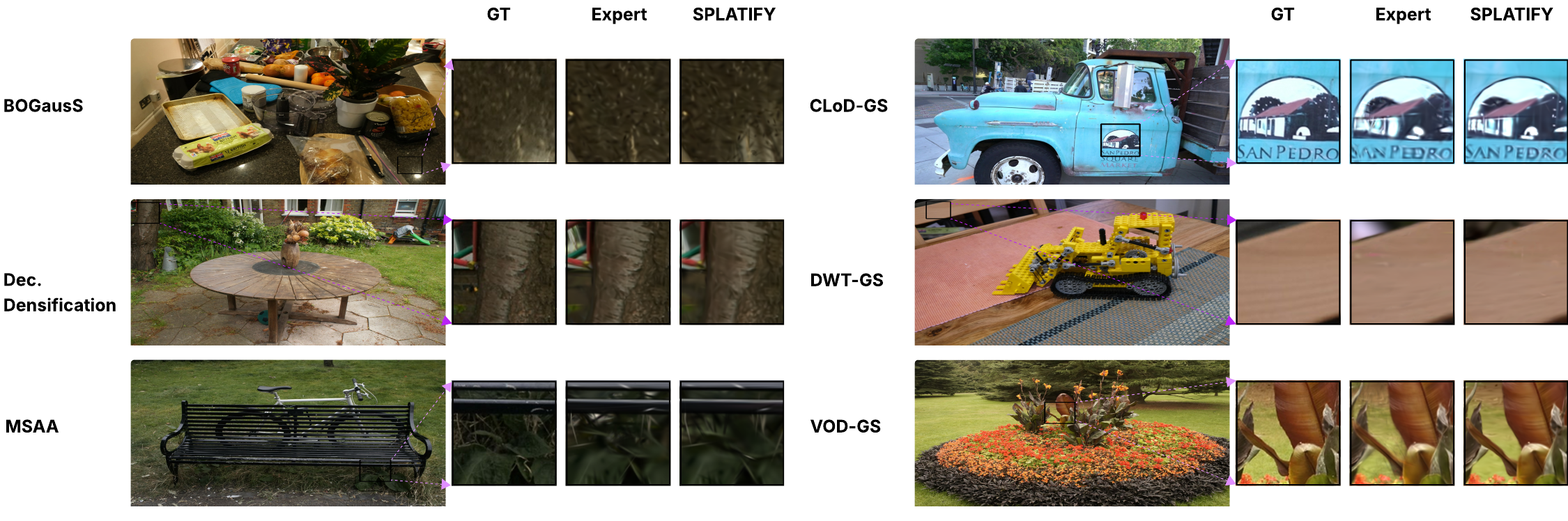}
  \caption{\textbf{Qualitative comparison of \splatify\ against baselines on Set~1 papers.} Rendered novel views from \splatify\ and expert human implementation. All baselines fail to produce trainable code; \splatify\ closely matches expert quality.}
  \label{fig:qualitative}
\end{figure}

\paragraph{Quantitative and Qualitative Results.}
Table~\ref{tab:comparison} shows results for representative Set~1 papers, each evaluated under its own reported protocol (most average across all nine Mip-NeRF~360 scenes; some report per-scene). \splatify\ achieves visual quality within $\pm$0.5~dB PSNR and $\pm$0.02 SSIM of expert implementations on average, with qualitative comparisons shown in Fig.~\ref{fig:qualitative}. All baselines fail to produce trainable code (Table~\ref{tab:executability}): generic systems cannot recover implicit dependencies from citations or structure multi-file gsplat plugins correctly.

\begin{table}[t]
\centering
\caption{\textbf{Executability comparison.} All baselines fail to produce trainable code. $\sim$ denotes occasional success that never reaches training stability or paper-level results.}
\label{tab:executability}
\resizebox{\columnwidth}{!}{
\begin{tabular}{lccccccc}
\toprule
\textbf{Metric} & \textbf{Paper2Code} & \textbf{AutoP2C} & \textbf{GPT-5.2} & \textbf{Gemini 3.1 Pro} & \textbf{DeepSeek 3.2}  & \textbf{NERFIFY} & \textbf{SPLATIFY (Ours)} \\
\midrule
Imports Resolve            & \checkmark & \texttimes & \checkmark & \checkmark & \checkmark & \checkmark & \checkmark \\
Compiles / Trainable       & \texttimes & \texttimes & $\sim$     & $\sim$     & \texttimes & \checkmark & \checkmark  \\
Training Stability         & \texttimes & \texttimes & \texttimes & \texttimes & \texttimes & \texttimes & \checkmark \\
Gaussian Invariants        & \texttimes & \texttimes & \texttimes & \texttimes & \texttimes  & \texttimes & \checkmark \\
Converges to Paper Results & \texttimes & \texttimes & \texttimes & \texttimes & \texttimes & $\sim$ & \checkmark \\
\bottomrule
\end{tabular}}
\end{table}

\subsubsection{Set 2: Papers with Public Code}
\label{sec:set2}
Table~\ref{tab:set2-comparison} compares \splatify\ against original author repositories under identical training conditions. \splatify\ achieves comparable performance with average deviations under 0.3~dB PSNR, while translating each method into a standardized gsplat plugin within the Nerfstudio ecosystem for fair cross-method comparison. Although LLMs may have encountered these codebases during pretraining, baselines still fail to produce trainable code even with this potential advantage.

\begin{table}[t]
  \caption{\textbf{Set~2: \splatify\ vs.\ original author repositories.} Metrics from original papers compared against \splatify's automated gsplat implementations trained under identical conditions. Full results in the supplementary.}
  \label{tab:set2-comparison}
  \centering
  \resizebox{\linewidth}{!}{
  \begin{tabular}{@{}llcccccccc@{}}
    \toprule
    \textbf{Paper} & \textbf{Scene} & \multicolumn{4}{c}{\textbf{Original Repository}} & \multicolumn{4}{c}{\textbf{\splatify\ (Ours)}}  \\ 
    \cmidrule(lr){3-6} \cmidrule(lr){7-10}
     & & PSNR$\uparrow$ & SSIM$\uparrow$ & LPIPS$\downarrow$ & \#GS ($10^6$)  & PSNR$\uparrow$ & SSIM$\uparrow$ & LPIPS$\downarrow$ & \#GS ($10^6$)  \\ 
    \midrule
    3DGS~\cite{kerbl20233dgs}                     & Bicycle & 25.246 & 0.771 & 0.205  & --    & 26.162 & 0.765 & 0.359 & 0.964 \\
    LP3DGS~\cite{zhang2024lp}                     & Bicycle & 24.906 & 0.737 & 0.264  & 2.51  & 24.639 & 0.740 & 0.189 & 1.310 \\
    GNS~\cite{deng2025naturalselection}                   & Truck   & 26.520 & 0.899 & 0.119  & 0.60  & 25.970 & 0.891 & 0.095 & 0.490 \\
    Micro-Splatting~\cite{lee2025microsplatting}   & Kitchen & 34.100 & 0.960 & 0.011  & 0.67  & 33.145 & 0.935 & 0.099 & 1.620 \\
    CompGS~\cite{navaneet2024compgs}              & Counter & 28.467 & 0.895 & 0.222  & 0.385 & 28.283 & 0.894 & 0.153 & 0.665 \\
    \bottomrule
  \end{tabular}
  }
\end{table}


\subsection{Novelty Coverage Analysis}
\label{sec:novelty-coverage}

A system that compiles and trains but silently drops a novel loss term has failed at the task even if aggregate metrics appear reasonable. We evaluate novelty preservation on 10 papers from Set~3 using four metrics following~\cite{jain2026nerfifymultiagentframeworkturning}: $C$ (fraction correctly implemented), $I$ (partial/incorrect), $M$ (missing), and hyperparameter fidelity $W = |\{n \in \mathcal{N} : |\theta_n - \hat{\theta}_n| < 0.1|\theta_n|\}| / |\mathcal{N}|$. We additionally compute a weighted semantic score $(\sum_i w_i s_i) / \sum_i w_i$, where $w_i \in [0,1]$ reflects component importance and $s_i \in \{0, 0.2, \ldots, 1.0\}$ rates completeness, recognizing functionally equivalent implementations while penalizing semantic errors. All scoring is performed by Claude Opus 4.6 as an LLM judge, provided with the source paper, generated codes, and a structured rubric; human experts with 3DGS development experience independently scored all components blind to system identity, agreeing with the judge at
$\kappa{=}0.87$ on the $C/I/M$ label and Spearman $\rho{=}0.92$ on the
semantic score, reading each implementation function-by-function against its paper and flagging the exact baseline deviations (e.g., Euler angles for quaternions, \texttt{Sigmoid} vs \texttt{Tanh} opacity) independent of PSNR; they also verified every component added in compositional discovery and cross-domain invention. \splatify\ attains near perfect scores with semantic scores above 0.94 across all papers (Table~\ref{tab:novelty-coverage}).

\begin{table*}[t]
  \caption{\textbf{Novelty coverage on Set~3.} $C$: correct, $I$: incorrect/partial, $M$: missing, $W$: hyperparameter fidelity, Score: weighted semantic score (0--1). Full results in the supplementary.}
  \label{tab:novelty-coverage}
  \centering
  \resizebox{\textwidth}{!}{
  \begin{tabular}{@{}l ccccc ccccc ccccc ccccc ccccc@{}}
    \toprule
    & \multicolumn{5}{c}{\textbf{Paper2Code}} 
    & \multicolumn{5}{c}{\textbf{Gemini 3.1 Pro}} 
    & \multicolumn{5}{c}{\textbf{DeepSeek 3.2}} 
    & \multicolumn{5}{c}{\textbf{GPT-5.2 Thinking}} 
    & \multicolumn{5}{c}{\textbf{\splatify}} \\
    \cmidrule(lr){2-6} \cmidrule(lr){7-11} \cmidrule(lr){12-16} \cmidrule(lr){17-21} \cmidrule(lr){22-26}
    \textbf{Paper}
      & $C${\footnotesize$\uparrow$} & $I${\footnotesize$\downarrow$} & $M${\footnotesize$\downarrow$} & $W${\footnotesize$\uparrow$} & Score{\footnotesize$\uparrow$}
      & $C${\footnotesize$\uparrow$} & $I${\footnotesize$\downarrow$} & $M${\footnotesize$\downarrow$} & $W${\footnotesize$\uparrow$} & Score{\footnotesize$\uparrow$}
      & $C${\footnotesize$\uparrow$} & $I${\footnotesize$\downarrow$} & $M${\footnotesize$\downarrow$} & $W${\footnotesize$\uparrow$} & Score{\footnotesize$\uparrow$}
      & $C${\footnotesize$\uparrow$} & $I${\footnotesize$\downarrow$} & $M${\footnotesize$\downarrow$} & $W${\footnotesize$\uparrow$} & Score{\footnotesize$\uparrow$}
      & $C${\footnotesize$\uparrow$} & $I${\footnotesize$\downarrow$} & $M${\footnotesize$\downarrow$} & $W${\footnotesize$\uparrow$} & Score{\footnotesize$\uparrow$} \\
    \midrule
TrickGS~\cite{armagan2025trickgs}   & 0.35 & 0.30 & 0.35 & 0.40 & 0.41 & 0.50 & 0.20 & 0.30 & 0.50 & 0.51 & 0.65 & 0.15 & 0.20 & 0.60 & 0.64 & 0.60 & 0.15 & 0.25 & 0.65 & 0.62 & 0.90 & 0.05 & 0.05 & 0.92 & 0.94 \\
2DGS~\cite{huang20242dgs}                 & 0.85 & 0.08 & 0.07 & 0.50 & 0.55 & 0.55 & 0.15 & 0.30 & 0.40 & 0.40 & 0.80 & 0.10 & 0.10 & 0.55 & 0.58 & 0.75 & 0.10 & 0.15 & 0.60 & 0.61 & 1.00 & 0.00 & 0.00 & 0.95 & 0.95 \\
Scaffold-GS~\cite{lu2024scaffoldgs}       & 0.35 & 0.20 & 0.45 & 0.60 & 0.34 & 0.45 & 0.10 & 0.45 & 0.45 & 0.44 & 0.40 & 0.35 & 0.25 & 0.40 & 0.42 & 0.60 & 0.25 & 0.15 & 0.55 & 0.59 & 1.00 & 0.00 & 0.00 & 1.00 & 1.00 \\
GaussianShader~\cite{jiang2024gaussianshader} & 0.45 & 0.30 & 0.25 & 0.40 & 0.40 & 0.55 & 0.28 & 0.17 & 0.45 & 0.50 & 0.60 & 0.22 & 0.18 & 0.48 & 0.58 & 0.72 & 0.15 & 0.13 & 0.62 & 0.73 & 1.00 & 0.00 & 0.00 & 0.95 & 1.00 \\
Pixel-GS~\cite{zhang2024pixelgs}    & 0.25 & 0.45 & 0.30 & 0.60 & 0.40 & 0.40 & 0.10 & 0.50 & 0.80 & 0.46 & 0.55 & 0.15 & 0.30 & 0.80 & 0.67 & 0.65 & 0.10 & 0.25 & 0.85 & 0.72 & 1.00 & 0.00 & 0.00 & 1.00 & 1.00 \\
\bottomrule
  \end{tabular}
  }
\end{table*}

\subsection{Knowledge Discovery: Beyond Reproduction}
\label{sec:knowledge-discovery}

We now evaluate \textsc{Splatify}'s ability to go beyond faithful reproduction in two ways: improving already-reproduced methods by composing complementary techniques from the knowledge base, and synthesizing entirely new methods for scene types that no published paper has addressed.

\subsubsection{Improving Existing Methods}
\label{sec:discovered-improvements}

For papers in Set~1 and Set~2, we run a second pass in which the agents query the knowledge base for techniques that address deficiencies identified during visual feedback. The agents receive no instructions about which techniques to apply; they must independently diagnose weaknesses, retrieve remedies, and validate that the composition improves quality. For comparison, we also ran the same loop with the explorer agents replaced by uniform random sampling over $\mathcal{S}$. Random sampling used 8 times as many tokens and reached only a quarter of the PSNR gain.


Table~\ref{tab:discovery} report results. Compositional discovery yields an improvement of up to +2.4 dB PSNR over the reproduction-only baseline. The goal is not to invent new primitives but to identify combinations of existing techniques that have not yet been explored; since the number of valid compositions grows quickly with published innovations, manual exploration of this space is infeasible.

\begin{table}[t]
\centering
\caption{\textbf{Compositional discovery on Set~1, Set~2 papers.} ``Before'' is the reproduction-only baseline; ``After'' is the best validated configuration found by the discovery pipeline. All numbers on same scene as Tables \ref{tab:comparison} and \ref{tab:set2-comparison}.}
\label{tab:discovery}
\scriptsize

\setlength{\tabcolsep}{2pt}
\fontsize{6}{7}\selectfont
\renewcommand{\arraystretch}{1.0}
\begin{tabularx}{\linewidth}{
@{}l
>{\centering\arraybackslash}p{0.055\linewidth}
>{\centering\arraybackslash}p{0.055\linewidth}
>{\centering\arraybackslash}p{0.055\linewidth}
>{\centering\arraybackslash}p{0.055\linewidth}
>{\centering\arraybackslash}p{0.055\linewidth}
>{\centering\arraybackslash}p{0.055\linewidth}
>{\centering\arraybackslash}p{0.055\linewidth}
>{\centering\arraybackslash}p{0.055\linewidth}
>{\raggedright\arraybackslash}X
@{}}
\toprule
 & \multicolumn{4}{c}{Before} & \multicolumn{4}{c}{After} & \\
\cmidrule(lr){2-5} \cmidrule(lr){6-9}
Paper & PSNR$\uparrow$ & SSIM$\uparrow$ & LPIPS$\downarrow$ & \#GS & PSNR$\uparrow$ & SSIM$\uparrow$ & LPIPS$\downarrow$ &  \#GS & Things Added \\
\midrule
Opti3DGS~\cite{farooq2025opti3dgs}
  & 27.71 & .841 & .209 & 1.40
  & 28.05 & .853 & .125 & 4.95
  & Adaptive densif.\ threshold + opacity entropy reg. \\
\addlinespace[1pt]
MSAA 3DGS~\cite{yan2024msaa3dgs}
  & 25.35 & .758 & .369 & 2.73
  & 26.19 & .791 & .293 & 2.96
  & Bilateral grid + losses \\
\addlinespace[1pt]
GNS~\cite{deng2025naturalselection}
  & 25.97 & .891 & .095 & 0.49
  & 26.15 & .902 & .097 & 0.61
  & AA rasterization + denser init + post-densif.\ losses \\
\addlinespace[1pt]
DWTGS~\cite{nguyen2025dwtgs}
  & 18.68 & .510 & .374 & 1.64
  & 21.09 & .725 & .281 & 2.43
  & Patch structure loss + opacity/normal/depth reg. \\
\addlinespace[1pt]
CompGS~\cite{navaneet2024compgs}
  & 28.28 & .894 & .153 & 0.67
  & 28.50 & .899 & .146 & 0.63
  & EMA centroid updates + grad-based codebook refine \\
\bottomrule
\end{tabularx}
\end{table}

\subsubsection{Cross-Domain Method Creation}
\label{sec:cross-domain}

\begin{figure}[t]
\centering
\includegraphics[width=\linewidth]{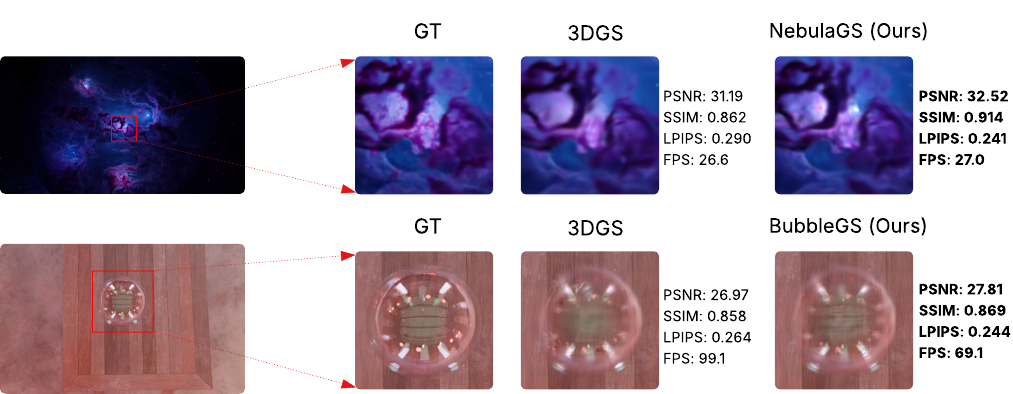}
\caption{\textbf{Cross-domain comparison.} Ground truth, 3DGS, and agent-generated method for Nebula and Bubble. NebulaGS resolves opaque shell artifacts; BubbleGS captures iridescent color variation. Additional scenes in the supplementary.}
\label{fig:crossdomain}
\end{figure}

All prior paper-to-code systems are limited to reproducing existing literature. We evaluate a qualitatively different capability: given a scene type that no published 3DGS paper has addressed, \textsc{Splatify} retrieves physical priors from outside the 3DGS community and composes them with rendering knowledge to synthesize a working method from scratch. We test on three domains where Gaussian primitives are a natural geometric fit, but standard 3DGS fails due to mismatched physical assumptions. Both scenarios are human-verified at the code level. Figure~\ref{fig:crossdomain} reports results.

\vspace{-0.5cm}
\paragraph{Scenario 1: Volumetric Nebula Rendering (NebulaGS).}
We evaluate on multi-view imagery of a planetary nebula (157 training, 23 test views). Nebulae are optically thin emission media whose intensity is a purely additive integral $I(\nu) = \int_0^{L} j_\nu(s)\,\mathrm{d}s$, with no absorption or occlusion. This directly conflicts with standard alpha compositing $C = \sum_i T_i\alpha_i c_i$ where $T_i = \prod_{j<i}(1-\alpha_j)$, which drives opacities toward $\{0,1\}$ and produces opaque shell artifacts. Nebular emission also follows Kolmogorov turbulence with energy spectrum $E(k) \propto k^{-5/3}$, placing significant detail at high spatial frequencies that pixel-space losses under-weight.

The agents autonomously retrieved this radiative transfer and turbulence theory and identified six modifications: (i)~an opacity ceiling loss ($\alpha < 0.95$) suppressing opaque shells; (ii)~a per-Gaussian emission parameter decoupling brightness from opacity; (iii)~a spectral frequency loss via 2D FFT log-amplitude matching; (iv)~a scale isotropy penalty encouraging spherical Gaussians; (v)~multi-scale L1 supervision to capture the Kolmogorov cascade; and (vi)~SH degree set to 0. All modifications are training-time losses requiring no rasterizer changes. NebulaGS achieves 32.52\,dB PSNR and 0.914 SSIM versus 31.19\,dB and 0.862 for vanilla 3DGS (+1.33\,dB, +0.052 SSIM), with per-view standard deviation dropping from 6.29 to 2.62 and rendering speed unaffected.
\vspace{-0.5cm}
\paragraph{Scenario 2: Iridescent Thin-Film Rendering (BubbleGS).}
Soap bubbles and oil slicks are dominated by thin-film interference. For a film of thickness $d$ and refractive index $n_f \approx 1.33$, the optical path difference is $\delta = 2n_f d\cos\theta_t$ where $\theta_t$ follows Snell's law $n_{\mathrm{air}}\sin\theta_i = n_f\sin\theta_t$. Per-wavelength reflectance follows the Airy function:
\begin{equation}
    R(\lambda) = \frac{2R_0(1-\cos\phi)}{1+R_0^2-2R_0\cos\phi}, \quad \phi = \frac{2\pi\delta}{\lambda},
    \label{eq:airy}
\end{equation}
where $R_0 = \bigl(\tfrac{n_f-1}{n_f+1}\bigr)^2 \approx 0.02$. The agents evaluate Eq.~\eqref{eq:airy} at $\lambda_R{=}650$\,nm, $\lambda_G{=}550$\,nm, $\lambda_B{=}450$\,nm and L2-normalize the result to obtain a physics-predicted unit chrominance $\hat{\mathbf{c}}_{\mathrm{phys}}$. Standard 3DGS fails because SH cannot represent the rapid color variation from Eq.~\eqref{eq:airy}, opacity binarization creates opaque shells where peak reflectance is only $4R_0 \approx 8\%$, no geometric prior encourages flat Gaussian arrangements.

The agents composed six modifications: (i)~a per-Gaussian film thickness parameter acting as soft segmentation; (ii)~a thin-film chrominance loss penalizing cosine distance between $\hat{\mathbf{c}}_{\mathrm{phys}}$ and rendered chrominance $\hat{\mathbf{c}}_{\mathrm{render}} = \mathbf{c}/\lVert\mathbf{c}\rVert_2$; (iii)~an opacity ceiling at $\alpha{=}0.45$ for film Gaussians; (iv)~a shell flatness loss encouraging disk-like Gaussians; (v)~total variation on the thickness map enforcing smooth drainage gradients; and (vi)~L1 sparsity on thickness. All losses are gated on the learned thickness and require no rasterizer changes. BubbleGS achieves 27.81\,dB PSNR and 0.869 SSIM versus 26.97\,dB and 0.858 for vanilla 3DGS, with LPIPS improving from 0.264 to 0.244 and per-view standard deviation dropping from 8.23 to 4.85. Rendering speed decreases from 99.1 to 69.1 FPS due to the second thickness rasterization pass, though both remain well above real-time. Both scenarios are human-verified at the code level. The supplementary video shows \splatify\ running end to end on a live paper, from PDF to trainable gsplat plugin, followed by a compositional discovery run.

\subsection{Ablation Study}
\label{sec:ablation}
\begin{table}[t]
\centering
\scriptsize
\setlength{\tabcolsep}{2pt}
\caption{\textbf{Component ablation.} Impact of each component on synthesis quality, averaged over 10 papers from \splatifybench.}

\begin{tabular}{lccc}
\toprule
\textbf{Configuration} & \textbf{Score} & \textbf{Train.} & \textbf{C} \\
 & & \textbf{(\%)} & \\
\midrule
\splatify\ (Full) & 0.97 & 100 & 1.00 \\
\midrule
\textit{Knowledge Sources:} \\
\quad w/o CFG Constraints (Stage 1) & 0.91 & 70 & 0.85 \\
\quad w/o In-context Examples (Stage 1) & 0.73 & 100 & 1.00 \\
\quad w/o Fork-aware Recovery (Stage 2) & 0.64 & 100 & 0.70 \\
\quad \hspace{2mm} w/ Simple Citation Recovery & 0.79 & 100 & 0.90 \\
\quad w/o Both (ICE + Recovery) & 0.52 & 70 & 0.65 \\
\midrule
\textit{Validation \& Feedback:} \\
\quad w/o Smoke Tests (Stage 3) & 0.68 & 50 & 1.00 \\
\quad w/o Gaussian Struct.\ Checks (Stage 3) & 0.89 & 80 & 0.95 \\
\quad w/o VLM Feedback (Stage 4) & 0.94 & 100 & 1.00 \\
\quad w/o PSNR Curve Monitor (Stage 4) & 0.94 & 100 & 0.95 \\
\midrule
\textit{Planning Strategy:} \\
\quad One-Shot w/o GoT (GPT 5.2 w/ other novelties) (Stage 3) & 0.49 & 60 & 0.95 \\
\bottomrule
\end{tabular}
\label{tab:ablation_all}
\end{table}

Table~\ref{tab:ablation_all} shows component ablations averaged over 10 Set~1 papers. CFG constraints have the largest impact on trainability (70\% without, vs.\ 100\% with), as agents misplace densification logic or wire optimizers incorrectly. Fork-aware recovery is equally critical: without it, 30\% of novelties go unimplemented. Smoke tests account for the largest single trainability drop (50\%), with Gaussian structural checks responsible for roughly half. PSNR curve monitoring catches late-stage failures invisible during short runs. One-shot generation without GoT collapses to score 0.49 despite high novelty fidelity, as agents produce monolithic files breaking dependency ordering.

\vspace{-0.3cm}
\section{Conclusion}

We presented \textsc{Splatify}, a multi-agent framework that converts 3DGS papers into trainable gsplat implementations through grammar-constrained synthesis, fork-aware citation recovery, and closed-loop visual feedback. On \textsc{Splatify-Bench}, the framework matches expert implementations within $\pm$0.5\,dB PSNR while all baselines fail to produce trainable code, reducing development time from weeks to minutes. Compositional discovery yields up to +2.4\,dB PSNR gain, and cross-domain invention produces working methods for scene types no prior paper has addressed, entirely without human guidance.

\noindent\textbf{Generalizations, limitations, and reproducibility.} The architectural elements of \textsc{Splatify} are not specific to Gaussian Splatting; adapting to a new domain such as 3D reconstruction pipelines or robotics simulation stacks requires defining a CFG over that framework's extension points and curating a seed set of verified paper-code pairs. Code, data, and all generated implementations will be publicly released.

{
    \small
    \bibliographystyle{ieeenat_fullname}
    \bibliography{main}
}


\end{document}